\documentclass[preprint,12pt]{elsarticle}

\usepackage{graphicx}
\usepackage{amssymb}
\usepackage{subcaption}

\usepackage{amsmath}
\usepackage{wrapfig}

\usepackage{lineno}

\journal{Artificial Intelligence in Medicine}

\begin{document}

\begin{frontmatter}


\title{Longitudinal modeling of MS patient trajectories improves predictions of disability progression}



\author[label2]{Edward De Brouwer\corref{cor}}
\ead{edward.debrouwer@esat.kuleuven.be}
\author[label3]{Thijs Becker\corref{cor}}
\ead{thijs.becker@uhasselt.be}
\author[label2]{Yves Moreau\corref{cor}}
\ead{moreau@esat.kuleuven.be}
\author[label5]{Eva Kubala Havrdova}
\author[label6]{Maria Trojano}
\author[label7]{Sara Eichau}
\author[label8]{Serkan Ozakbas}
\author[label9]{Marco Onofrj}
\author[label10]{Pierre Grammond}

\author[label11]{Jens Kuhle}
\author[label11]{Ludwig Kappos}
\author[label12]{Patrizia Sola}
\author[label13]{Elisabetta Cartechini}
\author[label14]{Jeannette Lechner-Scott}
\author[label15]{Raed Alroughani}
\author[label16]{Oliver Gerlach}
\author[label17,label18]{Tomas Kalincik}

\author[label19]{Franco Granella}
\author[label20]{Francois Grand’Maison}
\author[label21]{Roberto Bergamaschi}

\author[label22]{Maria José Sá}
\author[label23]{Bart Van Wijmeersch}
\author[label24]{Aysun Soysal}

\author[label25]{Jose Luis Sanchez-Menoyo}
\author[label26]{Claudio Solaro}
\author[label27]{Cavit Boz}
\author[label28]{Gerardo Iuliano}
\author[label29]{Katherine Buzzard}

\author[label30]{Eduardo Aguera-Morales}
\author[label31]{Murat Terzi}
\author[label32]{Tamara Castillo Trivio}
\author[label33]{Daniele Spitaleri}

\author[label34]{Vincent Van Pesch}
\author[label35]{Vahid Shaygannejad}
\author[label36]{Fraser Moore}
\author[label37]{Celia Oreja-Guevara}

\author[label38]{Davide Maimone}
\author[label39]{Riadh Gouider}
\author[label40]{Tunde Csepany}
\author[label41]{Cristina Ramo-Tello}
\author[label4,label3]{Liesbet Peeters\corref{cor}}
\ead{liesbet.peeters@uhasselt.be}

\address[label2]{ESAT-STADIUS, KU Leuven, 3001 Leuven, Belgium}
\cortext[cor]{Corresponding authors}

\address[label3]{I-Biostat, Data Science Institute, Hasselt University, Diepenbeek, Belgium}
\address[label4]{Department of Immunology, Biomedical Research Institute, Hasselt University, Diepenbeek, 3590, Belgium}
\address[label5]{Charles University in Prague and General University Hospital, Prague, Czech}
\address[label6]{Department of Basic Medical Sciences, Neuroscience and Sense Organs, University of Bari, Bari, Italy}
\address[label7]{Hospital Universitario Virgen Macarena, Sevilla, Spain}
\address[label8]{Dokuz Eylul University, Konak/Izmir, Turkey}
\address[label9]{University G. d’Annunzio, Chieti, Italy}
\address[label10]{CISSS Chaudire-Appalache, Levis, Canada}

\address[label11]{Neurologic Clinic and Policlinic, MS Center and Research Center for Clinical Neuroimmunology and Neuroscience Basel (RC2NB), University Hospital Basel, University of Basel, Basel, Switzerland}
\address[label12]{Azienda Ospedaliera Universitaria, Modena, Italy}
\address[label13]{Azienda Sanitaria Unica Regionale Marche - AV3, Macerata, Italy}
\address[label14]{University Newcastle, Newcastle, Australia}
\address[label15]{Amiri Hospital, Sharq, Kuwait}

\address[label16]{Zuyderland Ziekenhuis, Sittard, Netherlands}
\address[label17]{Melbourne MS Centre, Department of Neurology, Royal Melbourne Hospital, Melbourne, Australia}
\address[label18]{CORe, Department of Medicine, University of Melbourne, Melbourne, Australia}

\address[label19]{University of Parma, Parma, Italy}
\address[label20]{Neuro Rive-Sud, Quebec, Canada}
\address[label21]{IRCCS Mondino Foundation, Pavia, Italy}

\address[label22]{Department of Neurology, Centro Hospitalar Universitario de São João and University Fernando Pessoa, Porto, Portugal}
\address[label23]{Rehabilitation and MS-Centre Overpelt and Hasselt University, Hasselt, Belgium}
\address[label24]{Bakirkoy Education and Research Hospital for Psychiatric and Neurological Diseases, Istanbul, Turkey}

\address[label25]{Hospital de Galdakao-Usansolo, Galdakao, Spain}
\address[label26]{Dept of Rehabilitation mons L Novarese Hospital, Moncrivello, Italy}
\address[label27]{KTU Medical Faculty Farabi Hospital, Trabzon, Turkey}
\address[label28]{previously at Ospedali Riuniti di Salerno, Salerno, Italy}
\address[label29]{Box Hill Hospital, Melbourne, Australia}

\address[label30]{University Hospital Reina Sofia, Cordoba, Spain}
\address[label31]{19 Mayis University, Samsun, Turkey}
\address[label32]{Hospital Universitario Donostia, San Sebastain, Spain}
\address[label33]{Azienda Ospedaliera di Rilievo Nazionale San Giuseppe Moscati Avellino, Avellino,Italy}

\address[label34]{Cliniques Universitaires Saint-Luc, Brussels, Belgium}
\address[label35]{Isfahan Neurosciences Research Center, Isfahan University of Medical Sciences, Isfahan, Iran}

\address[label36]{Jewish General Hospital, Montreal, Canada}
\address[label37]{Hospital Clinico San Carlos, Madrid, Spain}
\address[label38]{Garibaldi Hospital, Catania, Italy}
\address[label39]{Razi Hospital, Manouba, Tunisia} 
\address[label40]{University of Debrecen, Debrecen, Hungary}
\address[label41]{Hospital Germans Trias i Pujol, Badalona, Spain}

\begin{abstract}
Research in Multiple Sclerosis (MS) has recently focused on extracting knowledge from real-world clinical data sources. This type of data is more abundant than data produced during clinical trials and potentially more informative about real-world clinical practice. However, this comes at the cost of less curated and controlled data sets. In this work, we address the task of optimally extracting information from longitudinal patient data in the real-world setting with a special focus on the sporadic sampling problem. Using the MSBase registry, we show that with machine learning methods suited for patient trajectories modeling, such as recurrent neural networks and tensor factorization, we can predict disability progression of patients in a two-year horizon with an ROC-AUC of $0.86$, which represents a 33\% decrease in the ranking pair error (1-AUC) compared to reference methods using static clinical features. Compared to the models available in the literature, this work uses the most complete patient history for MS disease progression prediction.
\end{abstract}


\end{frontmatter}


\section{Introduction}
\label{sec:intro}

Multiple Sclerosis (MS) is a chronic autoimmune disease characterized by heterogeneous progression across patients \cite{weiner2009challenge,mcfarland2007multiple}. This heterogeneity led to the clinical classification of different disease stages \cite{miller2007primary,confavreux2000relapses,lublin1996defining} with patients typically starting in the relapsing remitting (RR) phase, which can later progress to the secondary progressive (SP) phase. Clinical practice is aimed at keeping disability progression under control \cite{weiner2009challenge}. This led to the development of statistical methods to accurately predict the conversion from the relapsing remitting to the secondary progressive stage \cite{ion2017machine,zhao2017exploration}. However, to achieve more useful predictions, we would like to predict disease progression in a more detailed manner, for example using the Expanded Disability Status Scale (EDSS) \cite{kurtzke1983rating}. The EDSS is a score designed by clinicians to quantitatively assess patient disability with improved consistency and decreased subjectivity. This paper aims at predicting disability progression on the EDSS using longitudinal clinical patient data. This longitudinal data, referred to as 'patient trajectories' here, consists of the medical follow-up of patients over time along with the most important predictors such as current and past disability progression, past relapses, and most importantly current EDSS. Compared to previous approaches, which used mainly static information \cite{tousignant2019prediction,law2019machine}, using the detailed clinical history of each patient is expected to increase predictive power \cite{kalincik2017towards,ziemssen2016multiple,vrenken2013recommendations}.

One reason for not considering full patient trajectories in the past resides in the lack of data sets containing a large amount of patient-level longitudinal clinical data. Fortunately, advances in clinical practice and clinical data acquisition standards now facilitate the collection of large amounts of longitudinal data, both in terms of number of patients, but also in the number of clinical variables collected on a systematic basis. The MS community is particularly prolific in this regard with multiple international consortia, such as the MS Data Alliance \cite{peeters2020covid19}, MSBase \cite{trojano2017treatment,butzkueven2006msbase}, Multiple MS, or Big MS Network, as well as large registries, such as the Danish, Swedish and Italian registries \cite{koch1999danish,hillert2015swedish,trojano2019italian}.

However, complex longitudinal clinical data poses challenges for modeling. It is high dimensional, consists of different data types and is sparsely measured at a non-constant sampling rate. The non-constant sampling occurs because observations are only recorded at medical visits, which can be days, months, or even years apart. Clinicians may not perform all available tests at each visit. For instance, the number of hyperintense cerebral lesions on MRI are usually not available at each clinical visit. Suitable machine learning methods would therefore need to be able to optimally extract relevant information from this type of data. Common strategies for dealing with these challenges include imputation and time binning, which lead to loss of information and thus lower performance for the predictive task of interest.

In this work, we employ several models from the machine learning literature that can deal with sporadic time series and investigate their ability to predict disability progression of MS patients using their clinical trajectories. We study several model classes: Bayesian probabilistic tensor factorization (BPTF) \cite{simm2017macau}, continuous-time recurrent neural networks (RNN) \cite{neural_ode,de2019gru}, and time-aware recurrent neural networks \cite{baytas2017patient}. These models are trained on the task of predicting disability progression of individual patients over a 2-year horizon, achieving an ROC-AUC of $0.86$. We used one of the largest available MS registries, MSBase, to train and validate our models. To the best of our knowledge, this work uses the most complete patient history for MS disease progression prediction.
 
The structure of this paper is as follows. Section \ref{sec:related_works} presents related work that uses real-world patient trajectories, with an emphasis on MS. Section \ref{sec:materials} provides a detailed description of the task and of the patient cohort. Section \ref{sec:methods} describes the different models we propose, as well as the baselines we compare against. Sections \ref{sec:results} and \ref{sec:discussion} present the results of the methods we considered, their interpretations, and a vision for future work.

\section{Related work}
\label{sec:related_works}

Many recent publications have used statistical models and machine learning to distil new knowledge from MS real-world clinical data \cite{kalincik2017towards,cohen2020leveraging,trojano2017treatment}. Among them, some have developed methods using the longitudinal clinical history of the patients to predict or classify the disease course (more specifically the conversion from RRMS to SPMS) \cite{ion2017machine,zhao2017exploration,seccia2020considering}. With a different focus, Signori et al.~\cite{signori2018long} used patients disability trajectories to uncover patient subgroups using latent class mixed models, and showed that those groups had different probabilities of reaching an EDSS of 6.  

In contrast, our work aims at predicting disability progression, which is more specific given that patients with declining neurological capacity can remain in the same disease course category. There has been research focused on the prediction of the disability progression of MS patients, most of them using static features, and thus not considering the evolution of the patient over time. Among those, Tousignant et al.~\cite{tousignant2019prediction} used convolutional neural networks to predict prognosis from MRI scans from a single visit, achieving an ROC-AUC of around 0.70. Law et al.~\cite{law2019machine} proposed a decision tree approach based on static physiological variables. Yperman et al.~\cite{yperman2019machine} used random forests on features engineered on evoked potential time series to predict disability progression. Yet, to the best of our knowledge, there has been no work using longitudinal machine learning models to predict disability score progression from the full clinical history of MS patients.

\section{Materials and Methodology}
\label{sec:materials}

\subsection{Prediction task definition}

\begin{figure}[h]
  \includegraphics[width=\linewidth]{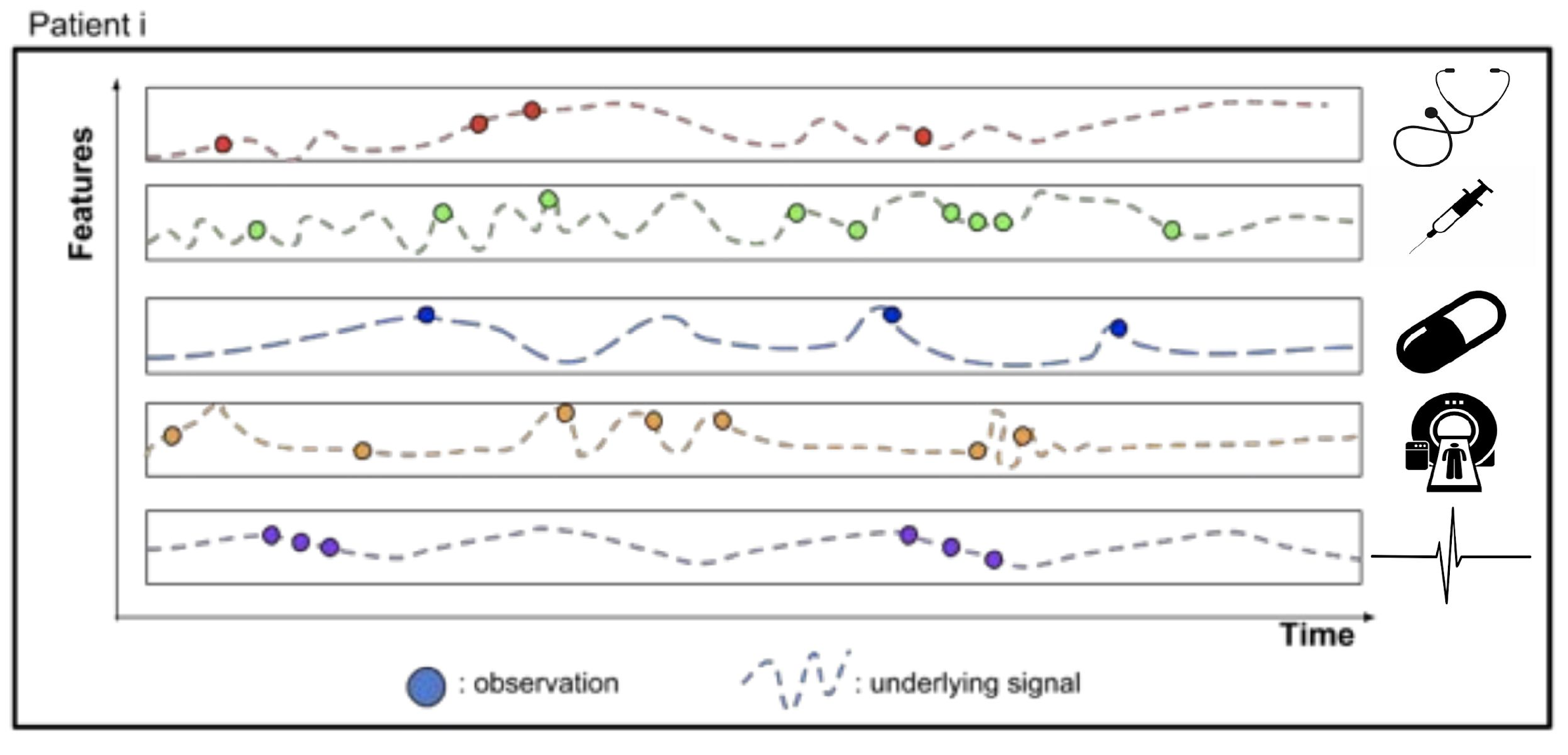}
  \caption{Illustration of sporadic time series for one patient. Dots stand for available measurements or observations while the dotted line stand for the true underlying process that would be observed in case of continuous follow-up. The sampling is very irregular in time as data is only collected during medical visits and all measurements are not sampled each time.}
  \label{fig:sporadic}
\end{figure}

We consider the task of predicting disability progression of patients based on their previous EDSS history. More formally, we have $N$ multiple sclerosis patients along with a matrix $X \in \mathbb{R}^{N \times d}$ of $d$-dimensional static covariates. For each patient $i$, we also have information about his or her medical history that we represent as a matrix $Y_i \in \mathbb{R}^{D \times N_{T_i}}$ and its corresponding vector of $N_{T_i}$ observations at times $t_i \in N_{T_i}$ where $D$ is the number of longitudinal variables. As every observation dimension might not be observed at every observation time, we also define a mask $M_i \in\{0,1\}^{D\times N_{T_i}}$.~If an observation is missing, the entry in the mask matrix and in $Y_i$ will be set to $0$. This configuration represents what we call a sporadic time series. The timing between observations varies from patient to patient: each has its own observation times $t_i$ and some observations might be missing at each observation time as more graphically represented on Figure \ref{fig:sporadic}.

\begin{figure}[h]
  \includegraphics[width=\linewidth]{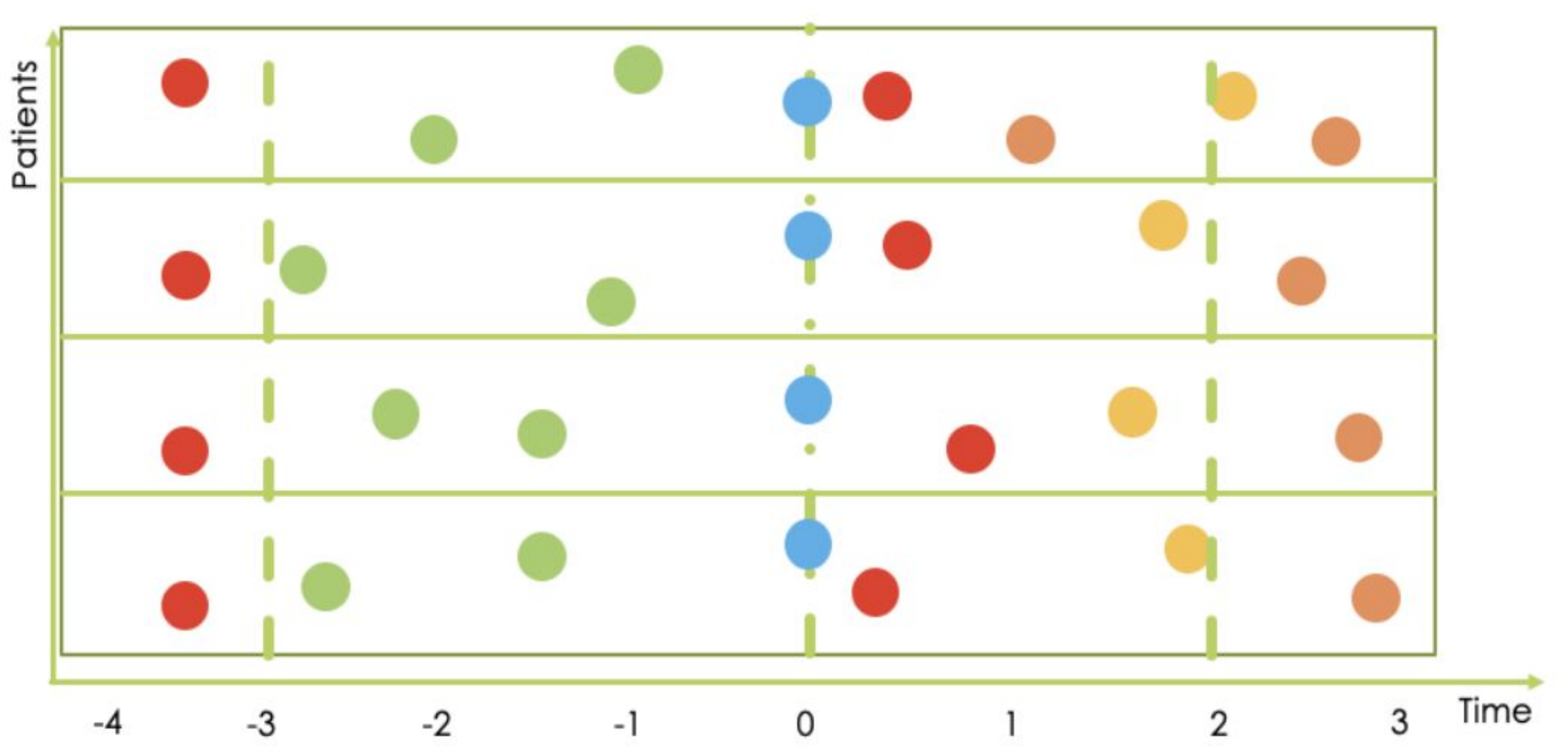}
  \caption{Graphical visualisation of the disability progression prediction task. Dots represent EDSS measurements over time. We aim at predicting disability progression at time $t=2$ from the information available at time $t \in [-3, 0]$  (\emph{i.e.}, we limit the EDSS progression history at 3 years back in time). The green and blue points represent the available EDSS measurements for prediction. Disability progression is defined with respect to the last observed EDSS in the observation window (blue). We define a $\Delta_{EDSS}$ as the difference between the EDSS closest to $t=2$ (orange) and EDSS at time 0 ($t=0$). Progression is assessed depending on the value of the last EDSS (blue) and the $\Delta$ (orange – blue) as in Equation \ref{eq:label_def}. Furthermore, only confirmed progressions are considered. That is, $\Delta_{EDSS}$ that are maintained over a period of at least 6 months. Note that we discard all EDSS measurements occurring less than 1 month after a relapse in the test period ($t>0$).}
  \label{fig:setup}
\end{figure}

Our goal is to use patient covariates $X_i$ and patient history $Y_i$ to predict disability progression after 2 years, based on the preceding 3-year trajectory. 
The binary label of disability progression $w$ after 2 years is defined as

\begin{align}
\label{eq:label_def}
    w_i &= \begin{cases}
    1 & \text{if $\Delta_{EDSS} \geq 1.5$  \&   $EDSS_{t_0} = 0$ } \\
    1 & \text{if $\Delta_{EDSS} \geq 1$  \&   $EDSS_{t_0} \leq 5.5$ }\\
    1 & \text{if $\Delta_{EDSS} \geq 0.5$  \&   $EDSS_{t_0} > 5.5$ }\\
    0 & \text{otherwise}
    \end{cases} \\
\end{align}

where $w=1$ indicates disability progression (also referred to as worsening). This 3-strata criterion is clinically motivated in \cite{kalincik2017towards,kalincik2015defining}. It takes into account that the EDSS scale is highly nonlinear (\emph{e.g.}, an increase of $1$ point over $5.5$ results in much higher impairment than for lower scores).
The time indexing of each patient starts at the observation time $t_0$ of the baseline $EDSS_{t_0}$, as illustrated on Figure \ref{fig:setup}. So $\Delta_{EDSS} = EDSS_{t_2^*}-EDSS_{t_0}$. The variables contained in $Y_i$ are those measured in the interval $t_i \in [-3,0]$. In practice, it rarely happens that another observation occurs exactly 2 years after $t=0$ so we refer to $EDSS_{t_2^*}$ as the closest observation from $t=2$, and occurring in the interval $t \in [1,3]$. Patients without at least one observation between $t=1$ and $t=3$ are therefore discarded.

To reliably assess disability progression, we use confirmed disability progression \cite{kalincik2017towards,kalincik2015defining}. We discard all EDSS measurements occurring less than 1 month after a relapse in the test period (\emph{i.e.,~}with $t>0$). Note that $EDSS_{t_2^*}$ can occur less than 1 month after a relapse. Progression should be confirmed by ensuring that all EDSS measurements for at least 6 months after $EDSS_{t_2^*}$ remain above the required threshold for disability progression as defined in Equation \ref{eq:label_def}. We require at least one confirmed EDSS measurement after $EDSS_{t_2^*}$.

Finally, we can define our task as predicting the worsening label $w_i$ from static data $X_i$ and historical data $Y_{i,t}$ where $t \in [-3,0]$.

\subsection{Cohort characteristics}

We used the cohort of MS patients from MSBase \cite{butzkueven2006msbase}, which contained at extraction time (August 2018) 53,687 unique patient records. We selected a subset of the initial cohort that complies with the following quality requirements. We first remove patients with missing or invalid diagnosis dates. This includes an invalid format or aberrant dates (dates in the future or before 1900). We remove all visit entries without EDSS value and with a date of visit before the onset date. We also removed all patients with visits before 1990 or with onset date before 1990. This is done with the motivation to analyze contemporary data and building a homogeneous patient cohort in terms of standard of care. 

We only selected patients with at least 6 visits in the three-year observation period (between $t=-3$ and $t=0$) so as to have enough extra information for the trajectories. This results in a loss of $5,367$ patients. 

After the cleaning procedure, we have $6,807$ patients among which $1,133$ patients progressed in disability after 2 years. Table \ref{tab:cohort} reports some summary statistics of the final cohort.
We received ethical approval for this study from the medical ethics committee of the University of Hasselt, number CME2019/059.

\begin{table}[h]
\centering
 \begin{tabular}{|l c c c c|} 
 \hline
 \textbf{Attribute} & \textbf{Mean [C.I.]} & \textbf{Std} & \textbf{Min} & \textbf{Max} \\
 \hline
 Total number of patients & 6807 & / & / & / \\
 EDSS counts per patient ($t\in [-3,0]$) & $9.37 [9.29,9.45]$ & $3.28$ & $6$ & $36$ \\ 
 Disease duration at $t=0$ [years] & $6.77 [6.67,6.88]$& $4.35$ & $3$ & $25$ \\
 Average EDSS per patient ($t\in [-3,0]$) & $2.35 [2.31,2.39]$ & $0.88$ & $0$ & $8.5$ \\
 Average EDSS per patient ($t \geq 0$) & $2.63 [2.59,2.67]$ & $1.07$ & $0$ & $9.1$ \\
 Age at onset [years] & $32.24 [32.02,32.46]$ & $9.25$ & $18$ & $73.46$ \\
 Female patients [\% of total] & 72.96\% & / & / & / \\
 CIS patients [total] & 407 & / &  $/$ & $/$ \\
 Primary Progressive patients [total] & $214$ & / & $/$ & $/$ \\
 Primary Relapsing patients [total] & $88$ & / & $/$ & $/$ \\
 Secondary Progressive patients [total] & $323$ & / & $/$ & $/$\\
 Relapsing-Remitting patients [total] & $5474$ & /  & $/$ & $/$ \\
 \hline
 \end{tabular}
 \caption{Summary statistics of the cohort of interest (mean, standard deviation, minimum value, and maximum value). For the mean, we provide the point estimate along with a 95\% confidence interval (CI).}
 \label{tab:cohort}
\end{table}

\section{Methods}
\label{sec:methods}

In this section, we define the modeling techniques used to meet the objective discussed in the previous section. We considered five models: a \emph{static} random forest trained on only the variables that are available at $t=0$, a \emph{dynamic} random forest trained on engineered features representing the patient trajectory between $t=-3$ and $t=0$, a Bayesian Probabilistic Matrix Factorization (BPMF) technique that can handle time series with missing data, a time-aware recurrent neural network, and GRU-ODE-Bayes, a continuous-time neural network model designed to deal with sporadic time series.

\subsection{Random forests}

Random forests are popular in the statistical and machine learning community as they are robust to overfitting and more interpretable than many other machine learning methods. In particular, they have been used extensively in the MS literature \cite{ion2017machine,seccia2020considering}. However, as mentioned in the introduction, those methods are not designed to take time series as input, especially if the time series is sporadic.

To overcome this problem, one usually simplifies the input data by extracting meaningful features and feeding them as a complete covariate vector to the random forest algorithm. More specifically, for each patient $i$, one extracts from $X_i$ and $Y_i$ some feature vector $z_i$ that is fully observed. That is, each dimension of $z_i$ can be computed for every patient. This fully observed vector can then be used along the target label $w_i$ to train a random forest model. 

The main difficulty in this type of approach is to extract informative features from the input data, which is also known as feature engineering. To highlight the information contained in the temporal medical history of the patients, we consider two sets of features: one static and one dynamic.

As their name suggests, the static feature set contains only static information about the patient, and thus nothing about temporal history, while the dynamic feature set contains information regarding the past clinical history. We now detail both feature sets.

\subsubsection{Static feature set}

In the static feature set, we ignore any past information about the patient. 
The features we retained in this setup are 

\begin{itemize}
\item Gender (binary)
\item Age at onset (in years)
\item MS course (stage of the disease the patient is currently in \cite{lublin1996defining} at time $t=0$: RRMS, SPMS, Primary Progressive MS (PPMS), or Clinically Isolated Syndrome (CIS))
\item Disease duration (years since onset at time $t=0$)
\item EDSS measured at that particular visit (at $t=0$)
\item Last used disease modifying therapies (DMT) at $t=0$.
\end{itemize}

Note that we included MS course and disease duration, which are actually representative of the patient clinical history. However, these are non-longitudinal variables, and they are generally available to the clinicians. For a complete description of the DMT groups used in the analysis, we refer the reader to~\ref{app:dmt}. 

\subsubsection{Dynamic feature set}
\label{sec:RF_dyn}
The dynamic feature set contains the static feature set and extends it with features that are meant to reflect information from the patient’s trajectory. As the history of the patient cannot be fed easily to the random forest, we have to select features that might contain relevant information in the trajectory. 

On top of previously listed features, the dynamic feature set includes

\begin{itemize}
    \item The EDSS closest to time $t=-3$, that is, the first EDSS that was measured for that patient.
    \item The maximum EDSS value that was reached over the observation window between $t=-3$ and $t=0$.
    \item The difference between the maximum and minimum EDSS in the observation window.
    \item The number of visits between $t=-3$ and $t=0$.
    \item The number of relapses in the observation period $t=-3$ and $t=0$.
\end{itemize}

Those 5 features are thought to be informative for the future course of the disease. Indeed, knowing EDSS at time $t=-3$ and $t=0$ gives us information about the slope of progression of the disease over 3 years. The maximum EDSS and the difference between the maximum and minimum contains information of the variability of the trajectory. 

\subsection{Bayesian tensor factorization}

The first method we addressed to deal with sporadically measured time series is Bayesian Probabilistic Tensor Factorization (BPTF), an extension of BPMF to tensors. In general, tensor factorization methods aim at approximating a tensor as the linear combination of $r$ rank-1 tensors \cite{mnih2008probabilistic}. To explain why we can use Bayesian factorization techniques here, we shall first give some details about the data representation.

\subsubsection{Data representation}

We stated in the previous section that each patient history was encoded in a matrix $Y_i$. By reworking this data representation and stacking all patient histories together, we can have a 3-mode tensor $\mathcal{Y}$. A 3-mode tensor has 3 axes and can be best thought of as a cube. In our case, the first axis would represent the patient index, the second the measurement type (here, we have only on measurement type: EDSS), and the third would be time such that $\mathcal{Y}_{i,j,t}$ would store the measurement type $j$ at time $t$ for patient $i$. This entails two main consequences. First, the time axis is shared for all the patients, meaning that some time binning will be needed and we will lose some temporal information. Second, most of the entries in the tensor will be empty (\emph{i.e.}, non-observed).

Binning the data in temporal bins leaves us with a trade-off. Small time steps would result in limited information loss but would make the tensor much sparser. To keep computations manageable without much temporal information loss, we chose a time bin of 30 days, which has also the advantage of being intuitive (1 month). With this binning factor, the tensor created with the data of our patients between $t=-3$ and $t=3$ has a filling rate of 21\% if we only consider EDSS. 

\subsubsection{Incorporating static features}

The patient trajectories can be encoded such as to be processed by BPTF. However, static features such as gender and disease course are very important for accurate prediction. To incorporate this source of information into the model, we consider two paths. The first is BPTF with side information as presented in \cite{simm2017macau}. However, this mapping is multilinear, which restricts the possibilities for the model to extract useful, possibly nonlinear, interactions between static covariates and the worsening label. To address this issue, we considered a second version of the model that consists of the same random forest model as described in Section \ref{sec:RF_dyn}, but where we extended the dynamic feature set with the prediction of the BPTF model at time $t=2$: $\mathcal{Y}_{i,j=EDSS,t=3}$. We call this variant BPTF-SI-RF. More technical details for both approaches are presented in \ref{app:BPTF}.

\subsection{Time-Aware Recurrent Neural Networks}

Standard recurrent neural network (RNN) architectures usually require a fixed step size in between observations, an assumption that this not met in the clinical time series we aim at analyzing. Yet, one can transform sporadic data into a sequence of observation vectors along with their observation times. For an observation matrix $Y_i \in \mathbb{R}^{D\times N_{T_i}}$ and time vector $t_i \in  \mathbb{R}^{N_{T_i}}$, we then build the sequence $Y^*_{i} \in \mathbb{R}^{(D+1)\times N_{T_i}}$  where the last row of $Y^*_{i}$ consists of the observation time. We can then feed this data representation to a recurrent neural network.
 
In this work, we consider a Gated Recurrent Unit (GRU) variant of RNNcell \cite{cho2014learning}. GRUs are lon- term memory cells and have the advantage of having fewer parameters than other options (\emph{e.g.~}Long Short Term Memory (LSTM)). We initialized the first hidden state of the GRU by feeding the static information through a multilayer perceptron (MLP) and compute the probability of worsening by feeding the last hidden state (after all observations have been processed) to another MLP. We call this model GRU-TA.

\subsection{GRU-ODE-Bayes}

The methods presented above used some artifice to deal with sporadic temporal data. The random forests use summary statistics of the trajectories, BPTF requires time binning of the time series, and time-aware RNNs consider time as if it were a feature. This has the obvious limitations of (1) losing data points (because of summarizing or of averaging in the binning case) and (2) degrading the timing accuracy of the measurements.

To more naturally accommodate the sporadic nature of the data, we use the GRU-ODE-Bayes model \cite{de2019gru}. GRU-ODE-Bayes was recently proposed as a new method to deal with sporadic time series. It assumes a continuous latent process $h(t)$ (some hidden health status) that generates the observations $Y(t)$ and tries to approximate the dynamics of the patient as shown on Figure 1. More technical details about the approach are presented in~\ref{app:gruode}.

\section{Results}
\label{sec:results}

\begin{figure}[h]
  \includegraphics[width=\linewidth]{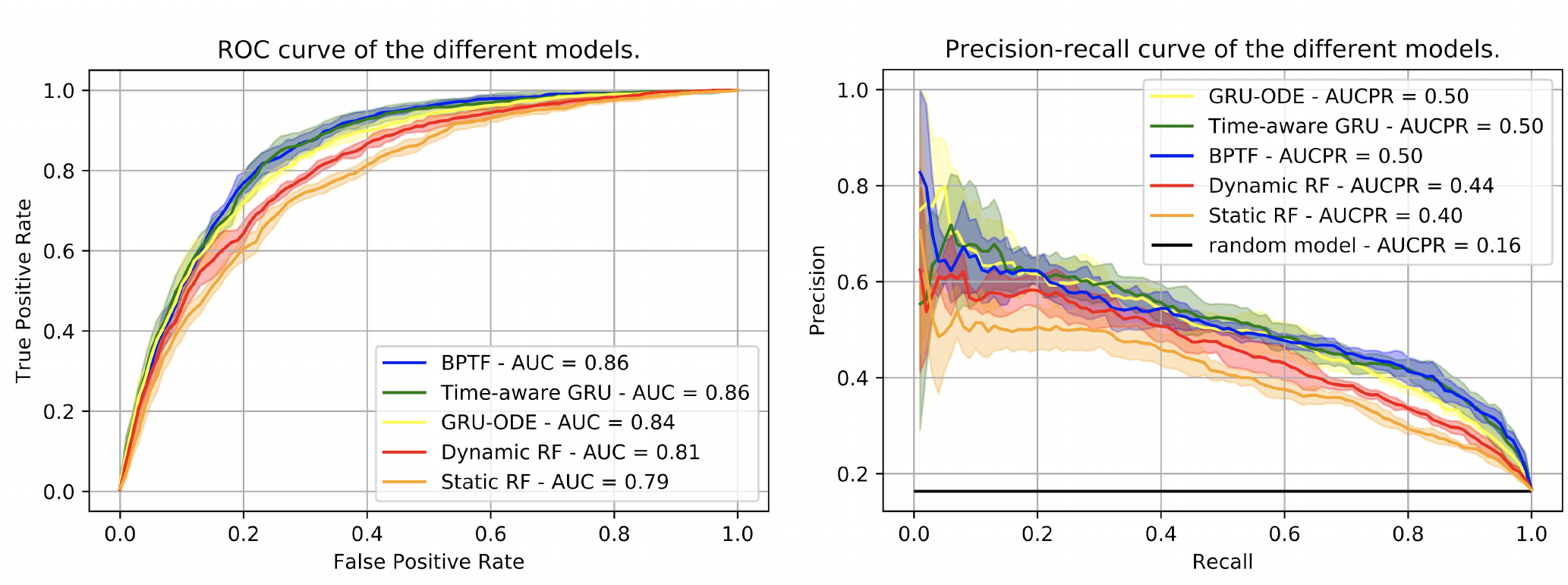}
  \caption{Receiver Operating Characteristic and Precision-Recall curves of the compared models.}
  \label{fig:roc_and_pr}
\end{figure}

To tune the hyperparameters, we used 5-fold cross-validation and used the exact same 5 training and validation sets over the different models for the sake of fair comparison. For the GRU-based models, we optimize the binary cross entropy. We report the average ROC-AUC and AUC-PR (precision-recall) metrics evaluated on 5 held-out test sets, as well as the standard deviation of those results. Performance results are displayed in Table \ref{tab:results}. We observe that static features only contain limited information for prediction of disability progression resulting in mediocre predictive performance (ROC-AUC of $0.79$ and AUC-PR of $0.40$ for the static feature set). Adding engineered temporal features improves the performances (ROC-AUC of $0.81$ and AUC-PR of $0.44$ for the dynamic feature set), but fails to harness as much information as models incorporating the full patient trajectories (ROC-AUC of $0.86$ for BPTF and GRU-TA and $0.84$ for GRU-ODE-Bayes, AUC-PR $0.5$). Figure \ref{fig:roc_and_pr} presents the ROC curve and the precision-recall curves of the compared models.

\begin{table}[h]
\centering
\begin{tabular}{l l c c}
\hline
\textbf{Model Type} & \textbf{Model Name} & \textbf{ROC-AUC} & \textbf{AUC-PR}\\
\hline
 & Random Model & $0.5$ & $0.16$ \\
Random Forest & Static feature set & $0.79 \pm 0.02$ & $0.40 \pm 0.02$ \\
Random Forest & Dynamic feature set & $0.81 \pm 0.01$ & $0.44 \pm 0.02$ \\

BPTF & BPTF-SI & $0.77 \pm 0.02$ & $0.42 \pm 0.02$ \\
BPTF & BPTF-SI-RF & $ \mathbf{0.86} \pm 0.01$ & $ \mathbf{0.50} \pm 0.02$ \\

Time-aware RNN & GRU-TA & $ \mathbf{0.86} \pm 0.01$ & $ \mathbf{0.50} \pm 0.03$ \\
ODE-RNN & GRU-ODE-Bayes & $\mathbf{0.84} \pm 0.01$ & $ \mathbf{0.50} \pm 0.03$ \\

\hline
\end{tabular}
\caption{Results for disability progression prediction with the different models. Best results are in bold. Several bolded values mean the results are not significantly different (significance assessed with pair-wise t-test).}
\label{tab:results}
\end{table}

\subsection{Patient trajectory analysis}

\begin{figure}

\centering

\begin{subfigure}{0.45\textwidth}
  \centering
  \includegraphics[width=0.95\linewidth]{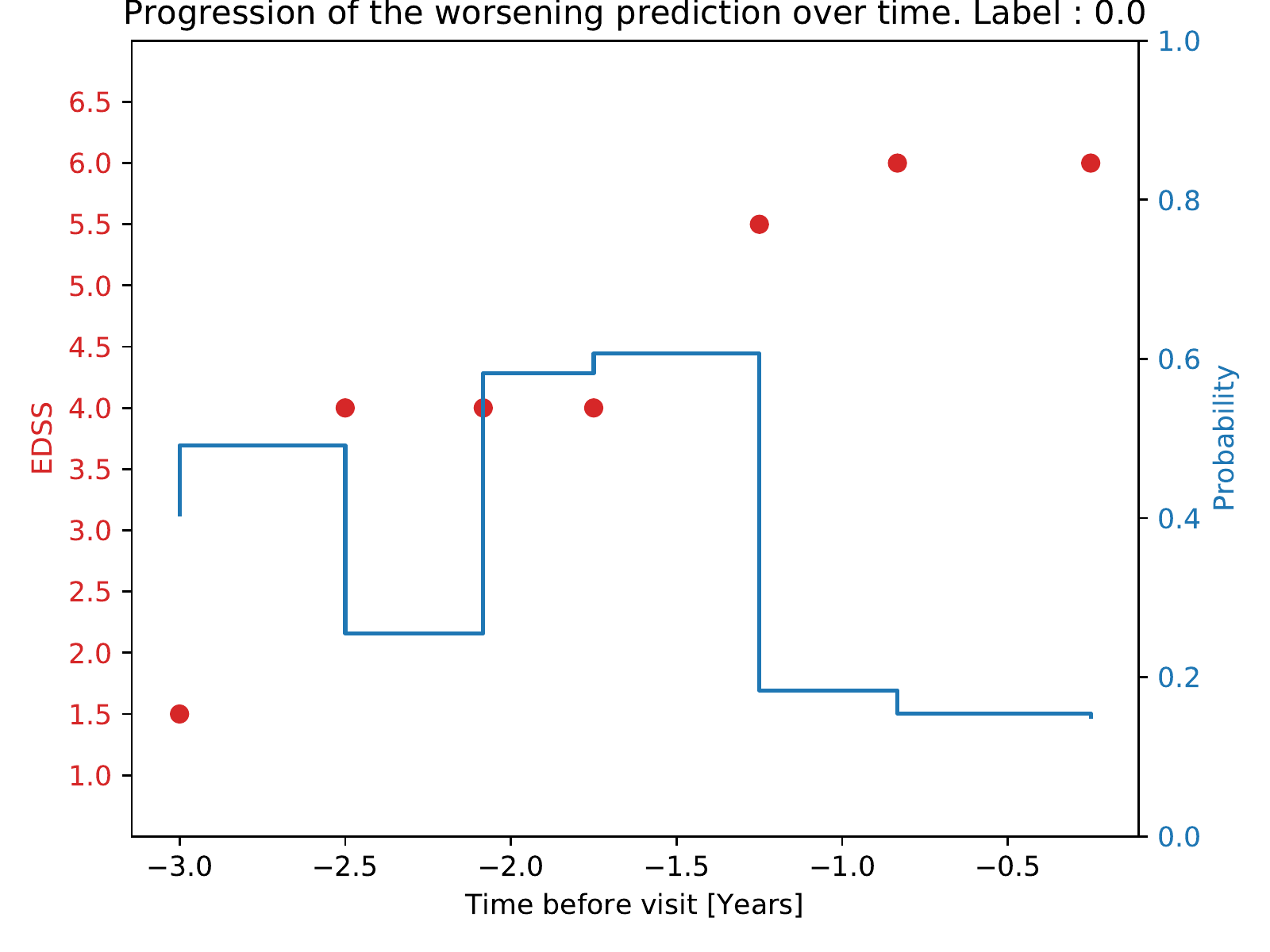}
  \label{fig:sub1}
\end{subfigure}%
\begin{subfigure}{0.45\textwidth}
  \centering
  \includegraphics[width=0.95\linewidth]{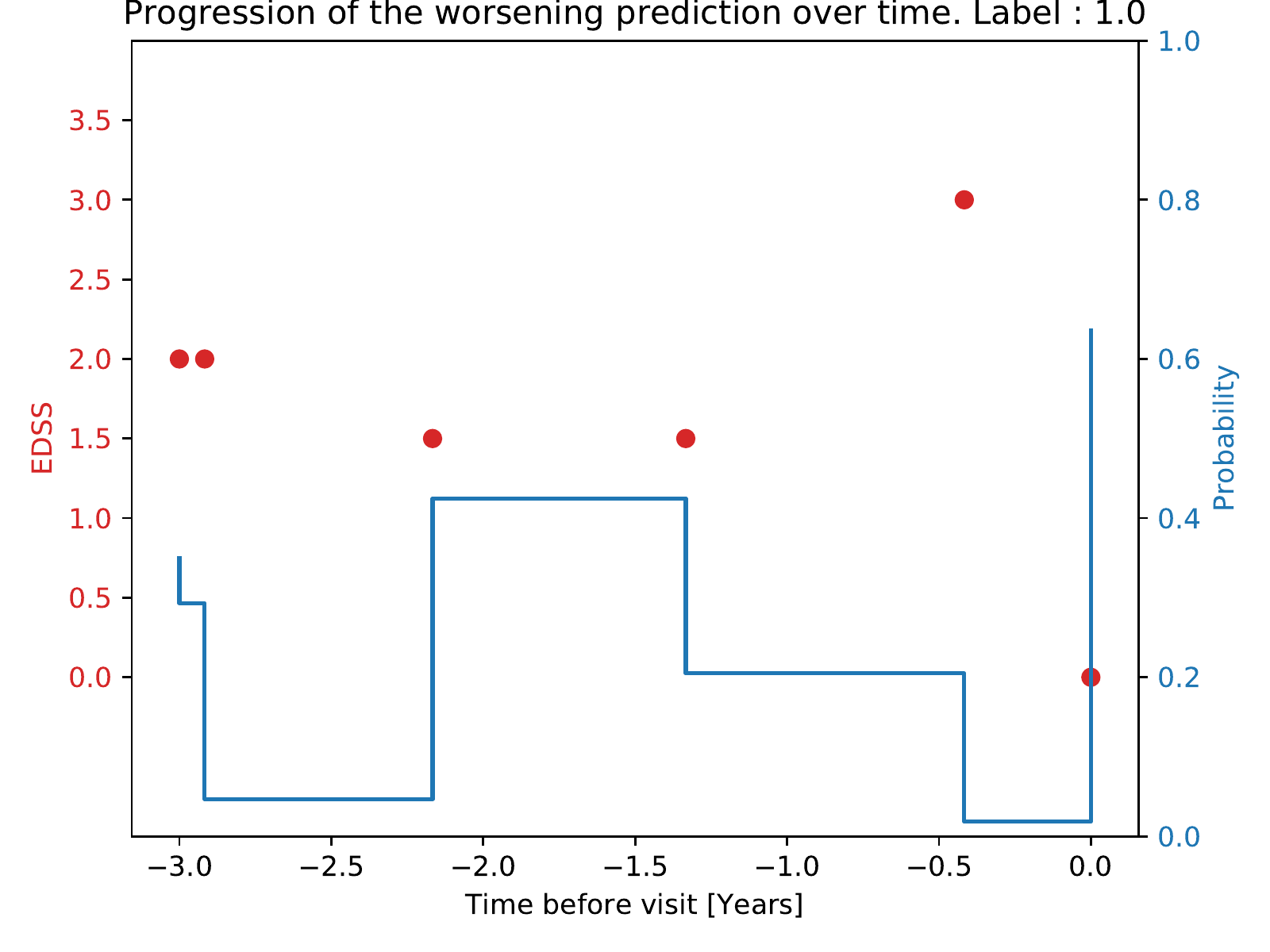}
  \label{fig:sub2}
\end{subfigure}

\medskip

\begin{subfigure}{0.45\textwidth}
  \centering
  \includegraphics[width=0.95\linewidth]{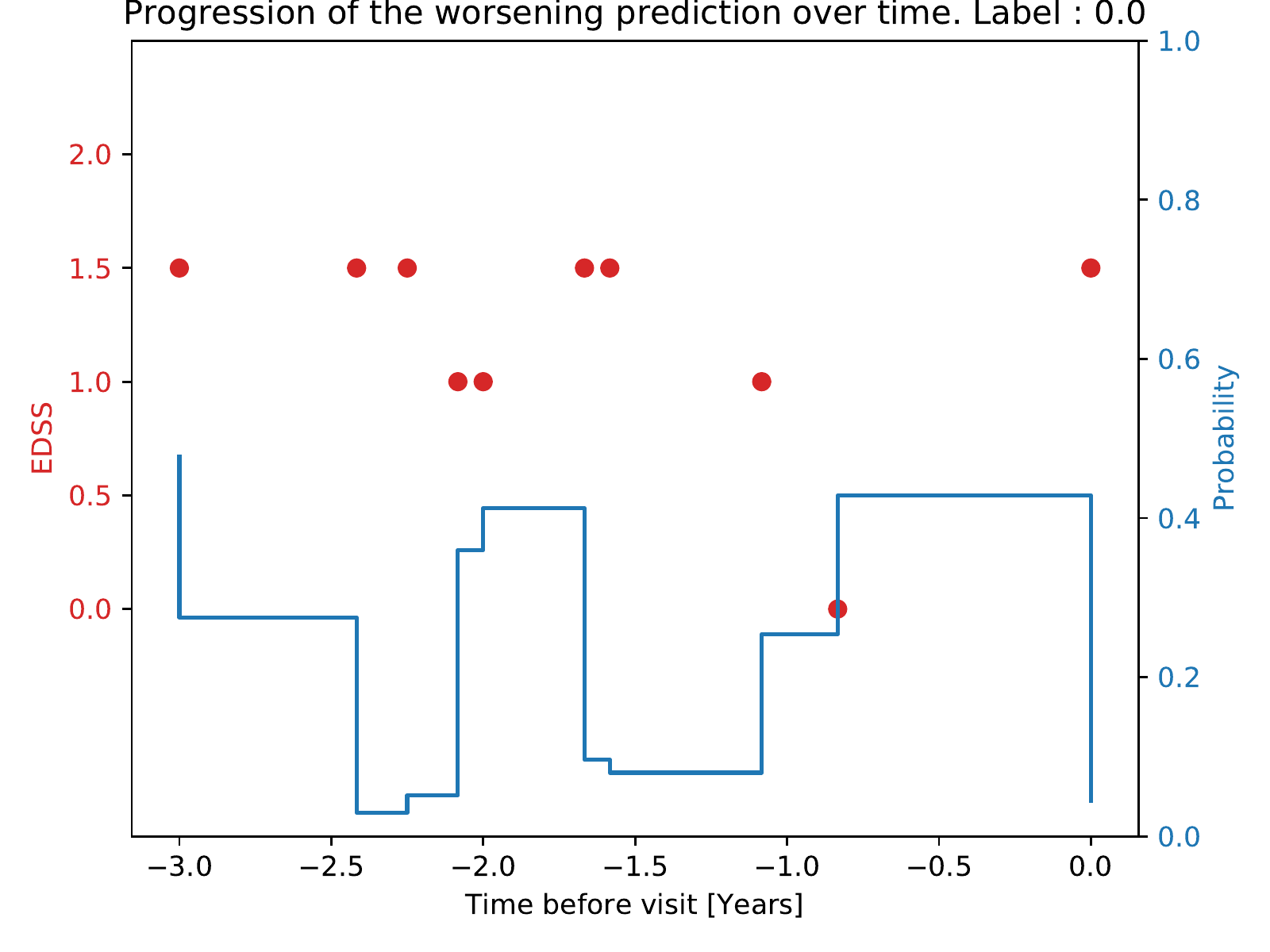}
  \caption{Non-worsening patients.}
  \label{fig:sub3}
\end{subfigure}%
\begin{subfigure}{0.45\textwidth}
  \centering
  \includegraphics[width=0.95\linewidth]{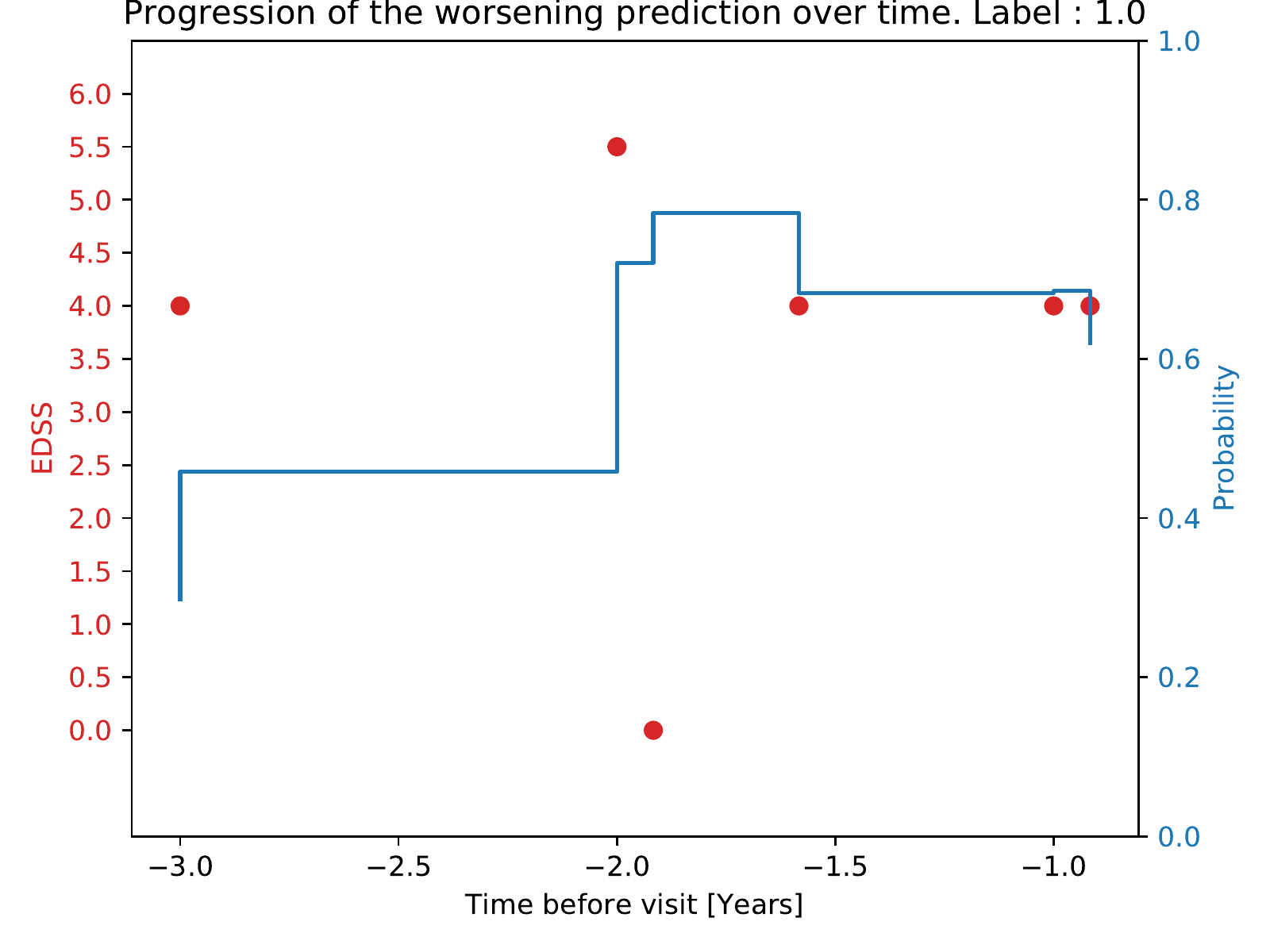}
  \caption{Worsening patients.}
  \label{fig:sub4}
\end{subfigure}

\caption{Evolution of the predicted probability of worsening for 2 sampled patients EDSS trajectories. Predictions are performed with GRU-ODE-Bayes. The blue line represents the probability of worsening at each point in time if no further measurement would be observed. The red dots stand for the observed EDSS values for that particular patient.}
\label{fig:trajectories}
\end{figure}

The GRU-ODE-Bayes model allows us to analyze the temporal evolution of the probability of worsening. Indeed, at each point in time, we can integrate the hidden process until $t=0$ and predict the worsening label. This allows us to evaluate the impact of the sequence of EDSS measurement on the worsening prediction. Figure \ref{fig:trajectories} shows an example of four chosen EDSS trajectories, two worsening and two non-worsening. At each point in time, we can compute the probability of worsening (from the GRU-ODE-Bayes model), would no other EDSS measurement be observed until the end of the observation period. The probability of worsening was calibrated with Platt scaling \cite{platt1999probabilistic}. On the left column of the figure (non-worsening patients), we observe that a probability of worsening seems to decrease when a significant EDSS progression is observed in the observation window suggesting a second progression 2 years after $t=0$ is less likely. On the right figure, for the worsening patients, we observe the same effect. When a significant drop in EDSS is observed, the model predicts that the EDSS is likely to rise again, to a level similar to the one it has achieved in the observation period. 

\subsection{Sensitivity analysis}

We performed a sensitivity analysis of the GRU-ODE-Bayes model, to assess the most predictive variables in the model. For each covariate, an importance score is determined as follows. The values of the covariate are shuffled randomly among the patients, making this covariate essentially non-predictive. The importance score is the average ROC-AUC degradation for the GRU-ODE-Bayes model, calculated from 10 repetitions (i.e., we repeat the shuffling of each covariate 10 times to improve the estimation of the average degradation effect). 

Table \ref{tab:sensitivity} presents the importance scores for the most impactful variables. The last observed EDSS is the most important feature. This result is not surprising as the task we aim at solving is to predict the worsening compared to the last observed EDSS. It is therefore predictable that this feature would be the most significant. Furthermore, we see that the full EDSS trajectory comes third in terms of feature importance. This confirms the importance of considering the whole patient clinical history for the prognosis. Other features are shown to be less important in the prediction of the worsening.

\begin{table}[h]
\centering
\begin{tabular}{l c}
\hline
\textbf{Feature} & \textbf{Sensitivity Score} \\
\hline
Last EDSS & $0.3 \pm 0.03$ \\
EDSS difference & $0.06 \pm 0.01$ \\

Full EDSS trajectory & $0.05 \pm 0.01$   \\
Max EDSS &  $0.03 \pm 0.01$  \\

Age at onset & $0.01 \pm 0.01$ \\
Others & $\leq 0.01$ \\

\hline
\end{tabular}
\caption{Sensitivity analysis of the features used in GRU-ODE-Bayes. Feature are presented by order of importance, together with their standard deviations.}
\label{tab:sensitivity}
\end{table}

\section{Discussion}
\label{sec:discussion}

The results above provide evidence that using the patient history results in more accurate prediction of future disease severity. Indeed, by adding simple summary features of the clinical history of the patients, we obtained an increase of performance of $0.02$ points of ROC-AUC and $0.04$ points of AUC-PR. Remarkably, this improvement was more pronounced when including the full EDSS trajectories in the modeling. In terms of ROC-AUC, BPTF (with a RF on top) and GRU-TA lead to the best performance with GRU-ODE-Bayes slightly below the other trajectory aware methods, but not significantly different (p-value $= 0.07$). This suggests the continuity of the latent health trajectory might not be fully satisfied here, with EDSS trajectories evolving in discrete updates (relapses) rather than continuously. This is in agreement with the findings of Lizak et al.~\cite{lizak2017highly}. Since GRU-ODE-Bayes assumes continuous underlying (latent) trajectories, the discrete nature of the relapses might hinder its performance. For AUC-PR, performances of all models using the full trajectories methods are on par.

The relevance of predictive models in clinical practice hinges on their capacity to detect with high precision all the patients that will experience a progression of the disease in the future. To correctly detect 80\% of progressing patients (recall$=0.8$), the static method would lead to a precision of 29\%. 71\% of the patients predicted as positive would then be false positives. With the full trajectory methods, the precision jumps to 42\%, which represents a clinically meaningful improvement in effect size in precision. To put things in perspective, let us consider a hypothetical cohort of 1,000 patients with similar statistics as the MSBase one. Out of those 1,000 patients, about 160 would eventually progress after 2 years. The static method could predict around 128 of those (80\%) and would wrongly detect approximately 313 patients as positives, which is about a third of the full cohort. The full trajectories method, on the other hand, would detect the same number of positives, but with a lower number of false positives: only about 176. This increase in precision leads to a more efficient clinical care as the limited resources of neurologists can be focused on a smaller and more specific subset of patients requiring special attention.

To assess quantitatively the information content in the full EDSS trajectories, we performed a sensitivity analysis with random permutations. It appeared that the last EDSS was crucial for prognosis. This is in line with clinical experience \cite{law2019machine,lizak2017highly}. Lizak \textit{et al.}~even suggested that disability evolution was an amnesic process at later stages of the disease \cite{lizak2017highly}. Yet the three other most important features are related to the patient clinical history (EDSS difference, full trajectory, max EDSS). In particular, removing the full EDSS trajectory resulted in an average large loss of 5 points of ROC-AUC. However, our predictions are averaged over the whole patient cohorts and we did not assess the impact on performance for a subset of more advanced disability patients only. 

Despite the quantitative evidence that taking past clinical trajectories into account for prognosis is beneficial, it is not yet clear which specific patterns are characteristics of future progression. From the example trajectories we provided in the results section, two main trends tend to appear. First, when a patient with initially stable EDSS experience some recovery (EDSS decreases), our model predicts a higher probability of worsening, suggesting the recovery is most probably temporary and the patient will progress over time.
Second, a patient with initially stable EDSS experiencing a progression during the observation window has a lower probability of future worsening, suggesting a patient having worsened significantly over the observation window is less likely to progress again afterwards. However, those interpretation are still qualitative and speculative and the design of dedicated methods to uncover specific patterns for prognosis of progression is left for future work.

\section{Conclusion}

In this study, we showed that including a more complete disability history of the patient in the statistical modeling improves the predictive performance of disability progression in MS. We considered several methodologies to incorporate those sporadic trajectories for the prediction of disability worsening of MS patients, eventually achieving state-of-the-art performance and showing quantitatively the impact of including the full EDSS trajectories in the modeling. This analysis confirms the importance of using longitudinal data to achieve AI-assisted precision medicine in MS. Indeed, we demonstrated an improvement of $13$ points of precision and $16$ points of specificity at a recall of 0.80, which translates into a more efficient stratification of patients to provide patients with optimal medical attention.

The evidence we provided in this paper suggests that more systematic collection of longitudinal patient data would be beneficial to patient followup, and that more accurate patient stratification and prognosis, based on the whole patient clinical history, will result in better and more patient-specific care in MS. One extra milestone towards this goal is to assess the efficacy of drugs more accurately than before (\emph{i.e.,} using the full trajectory to detect treatment response to a newly administered DMT). Cutting significantly the amount of time required for evaluating the effectiveness of a given treatment would result in lower disability progression during the optimal treatment search period. As treatment information (drug prescriptions) is available in the MSBase registry, we leave this temporal analysis of drug efficacy for future work.

\section*{Acknowledgements}

We would like to thank all patients and their cares who have participated in this study and who have contributed data to the MSBase cohort. The list of MSBase study group contributors are provided in \ref{app:authors}.

Yves Moreau is funded by Research Council KU Leuven: C14/18/092 SymBioSys3; CELSA-HIDUCTION CELSA/17/032
Flemish Government:IWT: Exaptation, PhD grants
FWO 06260 (Iterative and multi-level methods for Bayesian multirelational factorization with features). This research received funding from the Flemish Government under the “Onderzoeksprogramma Artificiële Intelligentie (AI) Vlaanderen” program. EU: “MELLODDY” This project has received funding from the Innovative Medicines Initiative 2 Joint Undertaking under grant agreement No 831472. This Joint Undertaking receives support from the European Union’s Horizon 2020 research and innovation program and EFPIA. Edward De Brouwer is funded by a FWO-SB grant.






\bibliographystyle{elsarticle-num-names}
\bibliography{sample.bib}

\vfill

\appendix

\section{MSBase investigators}
\label{app:authors}

This study would have been possible without the help of all MSBase investigators who contributed with patients data. By order of number of contributed patients : 

Eva Kubala Havrdova, Dana Horakova, Maria Trojano, Francesco Patti, Guillermo Izquierdo, Sara Eichau, Serkan Ozakbas, Marco Onofrj, Alexandre Prat, Marc Girard, Pierre Duquette, Pierre Grammond, Jens Kuhle, Ludwig Kappos, Patrizia Sola, Elisabetta Cartechini, Jeannette Lechner-Scott, Raed Alroughani, Oliver Gerlach, Tomas Kalincik, Franco Granella, Francois Grand'Maison, Roberto Bergamaschi, Maria Jose Sa, Bart Van Wijmeersch, Aysun Soysal, Ricardo Fernandez Bolaños, Jose Luis Sanchez-Menoyo, Claudio Solaro, Cavit Boz, Gerardo Iuliano, Katherine Buzzard, Olga Skibina, Julie Prevost, Eduardo Aguera-Morales, Murat Terzi, Tamara Castillo Triviño, Daniele Spitaleri, Maria Edite Rio, Vincent Van Pesch, Vahid Shaygannejad, Mark Slee, Fraser Moore, Celia Oreja-Guevara, Davide Maimone, Riadh Gouider, Tunde Csepany, Cristina Ramo-Tello, Edgardo Cristiano, Juan Ignacio Rojas, Shlomo Flechter, Maria Laura Saladino, Steve Vucic, Koen de Gans, Pamela McCombe, Radek Ampapa, Ayse Altintas, Norma Deri, Michael Barnett, Ernest Butler, Claudio Gobbi, Jose Antonio Cabrera-Gomez, Thor Petersen, Suzanne Hodgkinson, Richard Macdonell, Tatjana Petkovska-Boskova, Maria Pia Amato, Jose Andres Dominguez, Jabir Alkhaboori, Carlos Vrech, Guy Laureys, Gabor Lovas, Allan Kermode, Cameron Shaw, Anneke van der Walt, Helmut Butzkueven, Nikolaos Grigoriadis, Piroska Imre, Talal Al-Harbi, Neil Shuey, Angel Perez Sempere, Orla Gray, Magdolna Simo, Eniko Dobos, Cecilia Rajda, Bhim Singhal, Recai Turkoglu, Clara Chisari, Emanuele D'Amico, Lo Fermo Salvatore, Giovanna De Luca, Valeria Di Tommaso, Daniela Travaglini, Erika Pietrolongo, Maria di Ioia, Deborah Farina, Luca Mancinelli, Catherine Larochelle, Francesca Vitetta, Anna Maria Simone, Matteo Diamanti, Mark Marriott, Trevor Kilpatrick, John King, Katherine Buzzard, Ai-Lan Nguyen, Chris Dwyer, Mastura Monif, Izanne Roos, Lisa Taylor, Josephine Baker, Erica Curti, Elena Tsantes, Javier Olascoaga, Juan Ingacio Rojas and Freek Verheul.

\section{BPTF additional details}
\label{app:BPTF}
\subsection{Model details}

The Bayesian tensor factorization setting posits a specific multilinear generative model for the data. First, some latent matrices are generated from some prior for each of the modes of the tensor. Here, we have three : $U \in \mathbb{R}^{N \times r}$ for the patients axis, $V \in \mathbb{R}^{D \times r}$ for the measurements types axis and $W \in \mathbb{R}^{T_b \times r}$ for the time dimension where $T_b$ is the number of time bins between $t=-3$ and $t=3$. We consider the following generative process :

\begin{align}
U_{i,:} &\sim \mathcal{N}(\mu_a, \Sigma_a)  \text{ for } i \text{ in } 1...N \notag \\
V_{j,:} &\sim \mathcal{N}(\mu_b, \Sigma_b)  \text{ for } j \text{ in } 1...D \notag \\
W_{t,:} &\sim \mathcal{N}(\mu_c, \Sigma_c)  \text{ for } t \text{ in } 1...T_b \notag \\
\mathcal{Y}_{i,j,t} &\sim \mathcal{N}(\sum_{k=1}^K U_{i,k} V_{j,k} W_{t,k} , \alpha^{-1}),
\label{eq:TensDecomp}
\end{align}

where $\mu$ and $\Sigma$ stand for the means and covariance matrices of the prior distributions. The inference then consists in identifying the posterior probability distributions of the latent matrices $U$, $V$ and $W$ conditionally on the observed values of $\mathcal{Y}$ : 

\begin{align*}
    \mathbb{P}(U,V,W \mid \mathcal{Y}_{i,j,t} \text{ for each observed $(i,j,t)$ tuple})
\end{align*}

This inference is performed using Markov Chain Monte Carlo (MCMC) techniques and more specifically Gibbs sampling as shown in \cite{simm2017macau}. Once we have computed the posterior probability of the latent matrices $U, V$ and $W$, we can compute the posterior distribution of unseen samples, such as future EDDS values of patients. This can be computed using 

\begin{align*}
    & \mathbb{P}(Y_{i,j,t^*} = y \mid \mathcal{Y}_{i,j,t} \text{ for each observed $(i,j,t)$ tuple }) =  \\
    & \iiint  \mathbb{P}(Y_{i,j,t^*} y \mid U,V,W) \cdot \mathbb{P}(U,V,W \mid \mathcal{Y}_{i,j,t} \text{ for each observed $(i,j,t)$ tuple}) dU dV dW,
\end{align*}

where the first term is given by the generation model \ref{eq:TensDecomp} and second term is the posterior whose samples are generated by the MCMC routine.

\subsection{Adding side information}

To incorporate the static information we have about each patient, we use the Bayesian tensor factorization with side information framework. It consists in adding a linear mapping (a vector $\beta$ from the matrix of static covariates $X$ to the corresponding latents). In our case, we have such information for patients only, not for the other modalities. We then update the generation process as 

\begin{align*}
    \beta &\sim \mathcal{N}(\mu_{\beta}, \Sigma_{\beta}) \\
    U_{i,:} &\sim \mathcal{N}(\mu_a + \beta X_i, \Sigma_a)  \text{ for } i \text{ in } 1...N\\
\end{align*}

During inference, the posterior distribution of the vector $\beta$ will then have to be sampled as well. For this inference to be possible, the mapping between static covariates $X$ and the latents has to be linear. Would the mapping be more complex, it would quickly become intractable. For this reason, in practice, we merge the prediction of the BPTF with a random forest.

\section{GRU-ODE model}
\label{app:gruode}
\subsection{GRU-ODE}

The GRU-ODE module parametrizes the dynamics of the latent process $h(t)$ with an Neural-ODE inspired from the classical GRU module. We use the following parametric ODE as suggested in \cite{de2019gru}:
\begin{align}
  \label{eq:gru-ode-update}
  \dfrac{d h(t)}{dt} =  (1 - z(t)) \odot (g(t) - h(t)),
\end{align}
where  $\odot$ is the Hadamard product and $z(t)$ and $ g(t)$ are given as in the GRU equations: 

\begin{align}
  r_t &= \sigma(W_r x_t + U_r h_{t-1}+b_r) \nonumber \\
  z_t &= \sigma(W_z x_t + U_z h_{t-1}+b_z) 
  \label{eq:gru-original} \\
  g_t &= \tanh(W_h  x_t + U_h (r_t \odot h_{t-1}) +  b_h) \nonumber
\end{align}

\subsection{GRU-Bayes}

GRU-Bayes module is responsible for the update of the hidden state when new measurements are observed. As data comes in in packets, we allow the hidden process to jump to a new point in hidden space where it reflects more the newly observed data point.

This update is performed by using a GRU cell that takes as input the previous hidden state and the current observation and then mimics a Bayesian update to set the hidden to a new value that matches the current observations: 
\begin{equation}
    h(t_{+}) = GRU(h(t_{-}),
    f(\mathbf{y}[k], m[k], h(t_-)))
    \label{eq:GRU-Bayes}
\end{equation}

where $t_{-}$ and $t_{+}$ stand for the value of the vectors just before and after the update.

\subsubsection{All together}

At test time, we first compute the initial hidden state value $h(0)$ from the static covariates X with some neural network mapping $g(\cdot)$ :

\begin{align*}
    h(t=-3) = g(X).
\end{align*}

We integrate the hidden process according to the GRU-ODE dynamics until the first observation (done with numerical integration). When an observation is reached, we process it with GRU-Bayes and update the hidden state. We then resume to GRU-ODE integration from the new initial point and continue until a next observation is reached. At each point in time, we can use $f_{obs}(.)$ to predict the distribution of the measurements. When we run out of observations, the predictions over time are only performed by integrating GRU-ODE until the prediction time of interest. 

\section{DMTs used in the analysis}
\label{app:dmt}

We restricted the analysis to the following disease modifying therapies: 
\begin{itemize}
    \item Interferons
    \item Natalizumab
    \item Fingolimod
    \item Teriflunomide
    \item Dimethyl-Fumarate
    \item Glatiramer
    \item Alemtuzumab
    \item Rituximab
    \item Cladribine
    \item Ocrelizumab
    \item Other (contains stem cells therapy, Siponimod and Daclizumab)
\end{itemize}

If none of those DMTs were found in the history of patients (between $t=-3$ and $t=0$), the DMT field was set as \emph{no\_dmt\_found}.







\end{document}